# Curvature as a Baseline: Monocular Pose Measurement of Cylindrical Objects with Dual Colour Stripe Markers

Jaesung Hong[1,2] and Dong Won Lee[3]

[1] R&D Center, A4LAB Inc., South Korea

[2] Department of Robotics and Mechatronics Engineering, DGIST, South Korea

[3] Department of Otorhinolaryngology - Head and Neck Surgery, Daegu Catholic University, Daegu, Republic of Korea

Corresponding author: Jaesung Hong

Email: jhong@dgist.ac.kr; jhong@a4lab.co.kr

Co-author email: neck@cu.ac.kr (Dong Won Lee)

## Abstract

Thin cylindrical objects offer little space for planar markers. We propose a dual-colour-stripe marker that uses the known curvature of an object to improve pose measurement with a single camera. The boundary between two colours provides a long centre line, while dots and printed edges supply features at different depths. We analyse how this depth separation improves the estimation of rotation about the cylinder axis, or roll. For a camera-facing marker at 15 cm, the Cramér–Rao lower bound on roll standard deviation is 2.00° for a flat marker, 0.38° for a single curved marker and 0.21° for a wrap-around marker. Monte Carlo roll RMS errors are within about 10% of the corresponding standard-deviation bounds. A first-order relation connects the sagitta of the curved marker to the roll sensitivity provided by a stereo baseline. The analytical and simulation results explain how curvature improves monocular measurement and provide a basis for designing marker geometry.

## 1. Introduction

Many objects tracked in robotics and measurement systems have a cylindrical shape. Examples include probes, instrument shafts, drills and surgical cannulas. A printed fiducial marker offers a simple way to measure the six-degree-of-freedom pose of such an object. A calibrated camera detects known features, and a perspective-n-point (PnP) method estimates their position and orientation. Widely used markers such as ArUco [1] and AprilTag [2] normally use a flat square pattern. A narrow shaft provides little space for this pattern, and bending the print changes the geometry assumed by a planar pose estimator. Small planar markers also have a weakly constrained orientation near a frontal view, where two pose solutions can explain similar images [3], [4]. Infrared optical and electromagnetic trackers are established alternatives [5], [6], but require dedicated equipment and an attached tracking body or sensor.

A cylindrical surface can also provide useful measurement geometry. If its radius is known, the three-dimensional coordinates of features on the curved surface can be used for PnP calculation. Features across the curved surface then occupy different depths. A small rotation about the shaft changes their relative image positions, even when the marker faces the camera. This observation motivates our dual-colour-stripe (DCS) design. The two colours create a long straight boundary along the shaft. Dots within the stripes identify the pattern, and the printed boundaries add geometric information away from the centre line.

The main contribution is a quantitative account of how curvature changes the observability of a thin marker. We derive relations linking the sagitta of the curved print to near-frontal roll sensitivity and to an equivalent stereo baseline. Simulations test these relations and the predicted effects. The analysis explains the role of marker width, radius, viewing distance and roll angle. It therefore supports marker design for a specified object and working range, rather than treating curvature only as a deformation to be corrected. This provides a measurement principle for space-constrained optical metrology:

known surface curvature is linked to the local uncertainty of a single-camera roll estimate under a stated image-noise model. The resulting design guidance is relevant to researchers developing compact measurement systems for cylindrical probes and tools, where marker area and camera separation are limited.

A second contribution is the combined use of a long colour boundary and features at different depths. The centre line strongly constrains the projected shaft direction. Inner dots and outer edges constrain the remaining pose components through their known positions on the cylinder. Repeating the band around the circumference extends this principle to a wrap-around marker, whose band codes identify features across several visible surface regions.

The pose estimator also uses the predicted printed outline to refine an initial pose. It projects the marker model into the input image, measures the corresponding boundaries and updates the pose. This follows established model-based tracking principles [7], [8]. The contribution here is the choice and geometric interpretation of the observations supplied by the DCS pattern on a cylinder.

A simulation study evaluates the precision allowed by the feature geometry under a controlled noise model.

The theoretical bounds describe uncertainty caused by small image noise with correct feature correspondences. The simulation does not evaluate printing, calibration, detection or registration errors. This preprint presents a geometric principle and its numerical support; it does not report physical validation or establish clinical performance or absolute metrological accuracy.

## 2. Related work

Planar fiducial markers usually combine a recognisable pattern with geometric features for pose estimation. ARTag [9], ArUco [1] and AprilTag [2] use square patterns, and AprilTag 2 [10] improves detection efficiency. STag [7] adds a circular feature to improve stability. Comparative studies describe differences in detection and pose accuracy [11], and learned detectors extend their use to difficult lighting conditions [12]. ChromaTag [13] uses colour contrast to accelerate detection. In the proposed DCS marker, colour has an additional geometric role: the boundary between two stripes forms a long, clearly defined line that supports shaft-direction estimation. This allows the pattern to be elongated along the object instead of requiring a square mounting surface.

Several markers have been designed specifically for cylindrical objects. Zhang et al. [14] combined circular dots and chessboard vertices on surgical tools. Their method estimates candidate poses from coplanar dot centres and uses additional vertices outside that plane to resolve the ambiguous solution. CylinderTag [15] uses cyclic patterns encoded by projective invariants and reconstructs a three-dimensional feature model for pose estimation. PuzzlePoles [16] extends a combinatorial pattern around a pole to support identification from different directions. These studies establish that cylindrical markers and non-coplanar observations are useful. Our focus is the quantitative contribution of curvature to a thin colour-stripe marker, including which features supply the depth separation and how their contribution changes with viewing conditions.

Other approaches attach planar markers to three-dimensional carriers or combine them with additional sensing, as in recent near-infrared and RGB-D tracking systems [17], [18]. Such carriers provide useful feature geometry but add an external structure to the object. Our design follows the existing surface and can be applied as a flexible printed label without a bulky external support. Differences in marker dimensions, camera performance and evaluation procedures prevent a direct ranking of the reported accuracy.

## 3. Marker geometry

### 3.1. Dual colour stripes on a cylinder

The basic DCS marker is a 40 × 12 mm rectangle before attachment. Its 40 mm dimension runs along the shaft, and its 12 mm dimension runs around the circumference. These dimensions are an example and can be adapted to the object. Two colour stripes meet at the centre line. Each stripe contains three circular dots, 3 mm in diameter in this design, printed in the colour of the opposite stripe. The examples use blue/orange, magenta/green and red/cyan pairs. Their colours distinguish markers, while the asymmetric dot spacing identifies feature order. Marker dimensions and colours can be adapted to the object, provided that the camera can resolve the features and distinguish the two colours. Figure 1 shows a single-band design for flat and curved markers and a four-band wrap-around design.

The model uses the radius of the printed marker surface. The reference design has a nominal shaft diameter of 15 mm and an effective print-surface radius of 7.64 mm. The 12 mm band spans approximately 90° at the effective radius. Let u denote distance across the marker print, measured along the circumferential arc from the centre line, and let z denote distance along the shaft. Any printed feature point is mapped to the cylinder by

$$P(u,z)=\left[R\sin(u/R),-R(1-\cos(u/R)),z\right]^{T}, \quad (1)$$

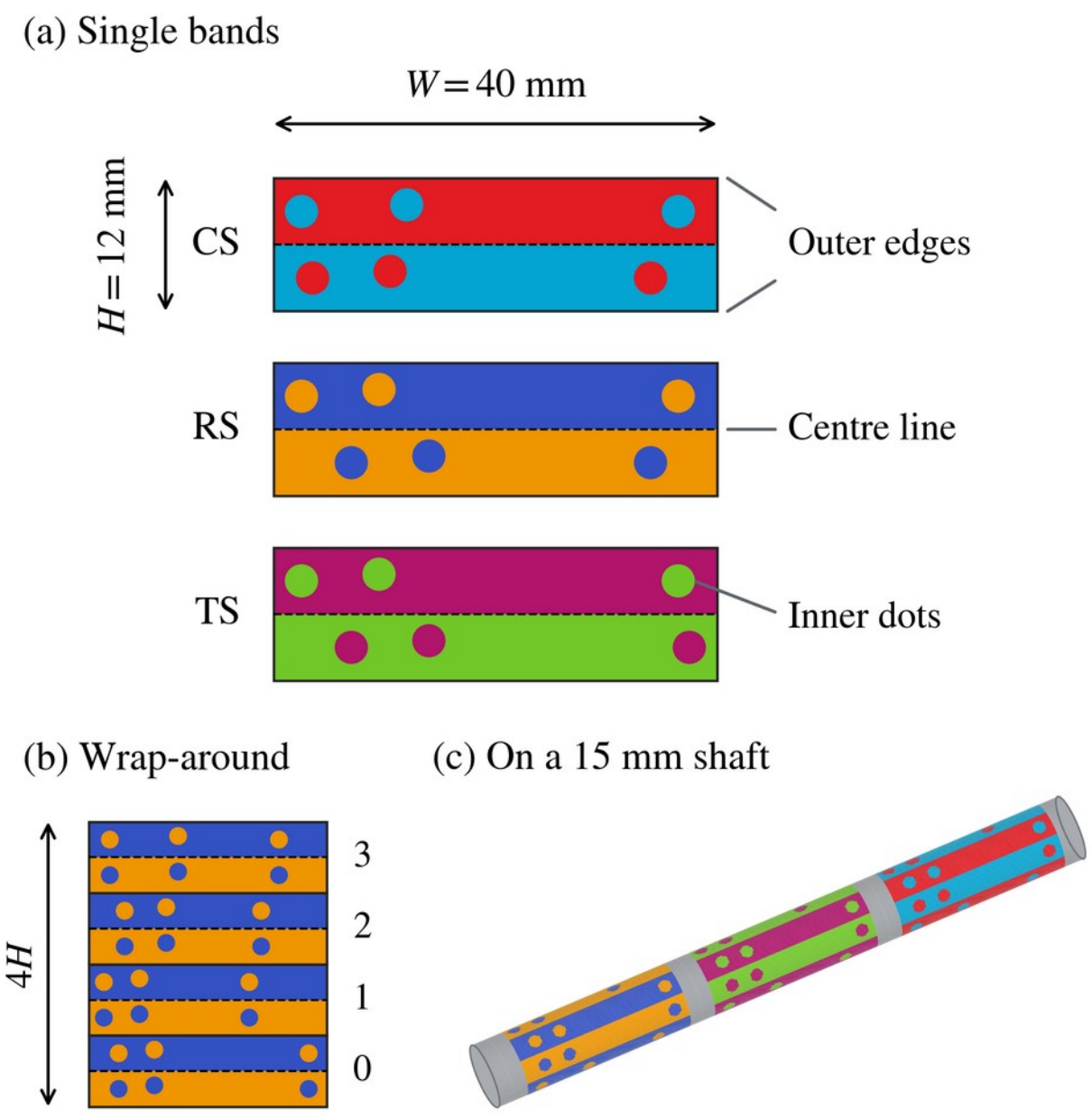

Fig. 1. DCS marker family. (a) Single bands CS, RS and TS for flat or curved markers, each 40 × 12 mm. (b) Four coded bands form a 40 × 48 mm wrap-around marker. (c) Schematic view of wrap-around markers on a nominal 15 mm shaft. The effective print-surface radius is 7.64 mm.

The origin is the midpoint of the centre line on the print surface. The x axis is tangent to the surface, the y axis points outward from the cylinder axis, and the z axis follows the shaft. The cylinder axis passes through $[0, -R, 0]^T$. A flat marker uses $P = [u, 0, z]^T$. The curved model retains the depth differences that a planar approximation would remove. At either outer edge, this difference from the centre-line tangent plane is the sagitta s:

$$s=R\left[1-\cos(H/2R)\right]\approx 2.24\,\text{mm} \quad (2)$$

For H = 12 mm and R = 7.64 mm, s is about 2.24 mm. Figure 2 shows how the centre line, dots and outer edges occupy different depths. The centre line provides a long observation of the projected shaft direction. The dots provide identifiable point correspondences and axial spacing. The outer edges add observations farther from the centre line, where the depth separation is greater.

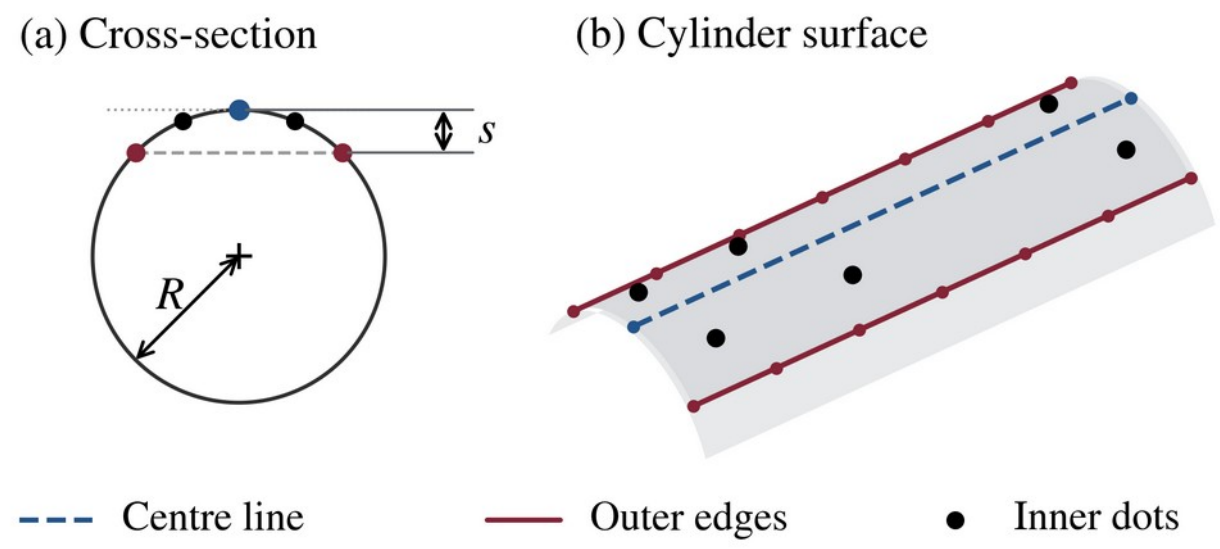

Fig. 2. Features of a curved band. (a) Cross-section showing the sagitta s between the centre-line tangent plane and the outer edges. (b) Centre line, six inner dots and sampled outer edges on the cylindrical surface. Both outer edges run parallel to the cylinder axis.

### 3.2. Wrap-around marker

A single curved band has a limited range of full visibility. Under a distant-view approximation, the 90° band remains fully visible for roll angles of about ±45° from the frontal view. Beyond this range, some features disappear around the shaft. A wrap-around marker places several coded bands around the circumference so that another band becomes visible as the object turns. Here four 12 mm bands form a 48 mm circumference, consistent with an effective radius of approximately 7.64 mm. Each band has a distinct dot-spacing pattern. Decoding the band assigns its dots to known circumferential positions, allowing visible dots from adjacent bands to contribute to one pose estimate. The depth separation across these bands can be larger than that within a single band. The two axial ends of the strip also supply boundary observations, serving a similar role to the outer edges of a single curved band. The outward normal of band 0 defines the reference for absolute roll.

## 4. Geometric basis of pose precision

The three-dimensional feature coordinates determine how a pose change appears in the image. We examine the weak roll information of a thin flat marker, the additional information supplied by curvature, and the role of the centre line. The analysis uses local changes about a known pose. It explains sensitivity and uncertainty near that pose, rather than guaranteeing a unique solution for every image.

### 4.1. Planar pose ambiguity

Planar pose ambiguity is an established problem [3], [4]. Three related difficulties are important for a planar marker near a frontal view: distinguishing opposite roll signs (the flip ambiguity), separating roll-induced image motion from translation, and separating the apparent narrowing caused by tilt from that caused by increased distance. The full feature pattern can distinguish these cases, but the differences may be small relative to image noise when perspective effects are weak. Depth-separated observations provide additional cues to distinguish the poses. Zhang et al. [14] use out-of-plane features to select a pose; the proposed DCS marker incorporates such depth differences directly in pose estimation. Curvature notably reduces these ambiguities when the relevant features are visible and correctly identified.

### 4.2. Roll sensitivity from curvature

Consider a thin flat print with half-width $h$ and roll angle $\psi$. Under weak perspective, its projected half-width is proportional to $h \cos \psi$. The rate of change is proportional to $-h \sin \psi$ and is zero at the frontal pose. A small roll therefore changes the apparent width very little near $\psi = 0$. For a curved print, the outer edges lie a distance $s$ below the tangent plane at the centre line. A small roll $\delta\psi$ produces a relative lateral motion of approximately $s\,\delta\psi$ between these edges and the centre line. After projection, the corresponding image motion is approximately $(f/\bar{Z})s\,\delta\psi$, where $f$ is focal length in pixels and $\bar{Z}$ is working distance. This relative motion remains first order at the frontal pose. Figure 3 compares these relative image changes after alignment of the centre line.

A frontal flat marker is not completely insensitive to roll. Under the full pinhole model, the two points at ±h move in opposite depth directions as the marker rotates. Let m denote their projected midpoint relative to the projected centre line. After removing the common translation, its flat-marker sensitivity and the leading curvature contribution are

$$\frac{\partial m_{flat}}{\partial \psi}\Big|_{\psi = 0} = \frac{fh^2}{\bar{Z}^2}, \frac{\partial m_{curved}}{\partial \psi}\Big|_{\psi = 0} \approx \frac{fs}{\bar{Z}} \quad (3)$$

The flat-marker term is nonzero, but very weak. It decreases with the square of distance, whereas the curvature term decreases with distance itself. For the present band, the curvature term is about 6–16 times larger over a 10–25 cm range. Part of the flat-marker midpoint shift can also be explained by translation, which is another unknown in a six-parameter pose estimate. Curvature instead creates a differential shift between features at different depths. Under weak perspective, a common translation cannot reproduce that differential shift. This distinction explains the improved roll precision near a frontal view. The curved expression retains the leading term proportional to $s/\bar{Z}$; smaller perspective terms are omitted.

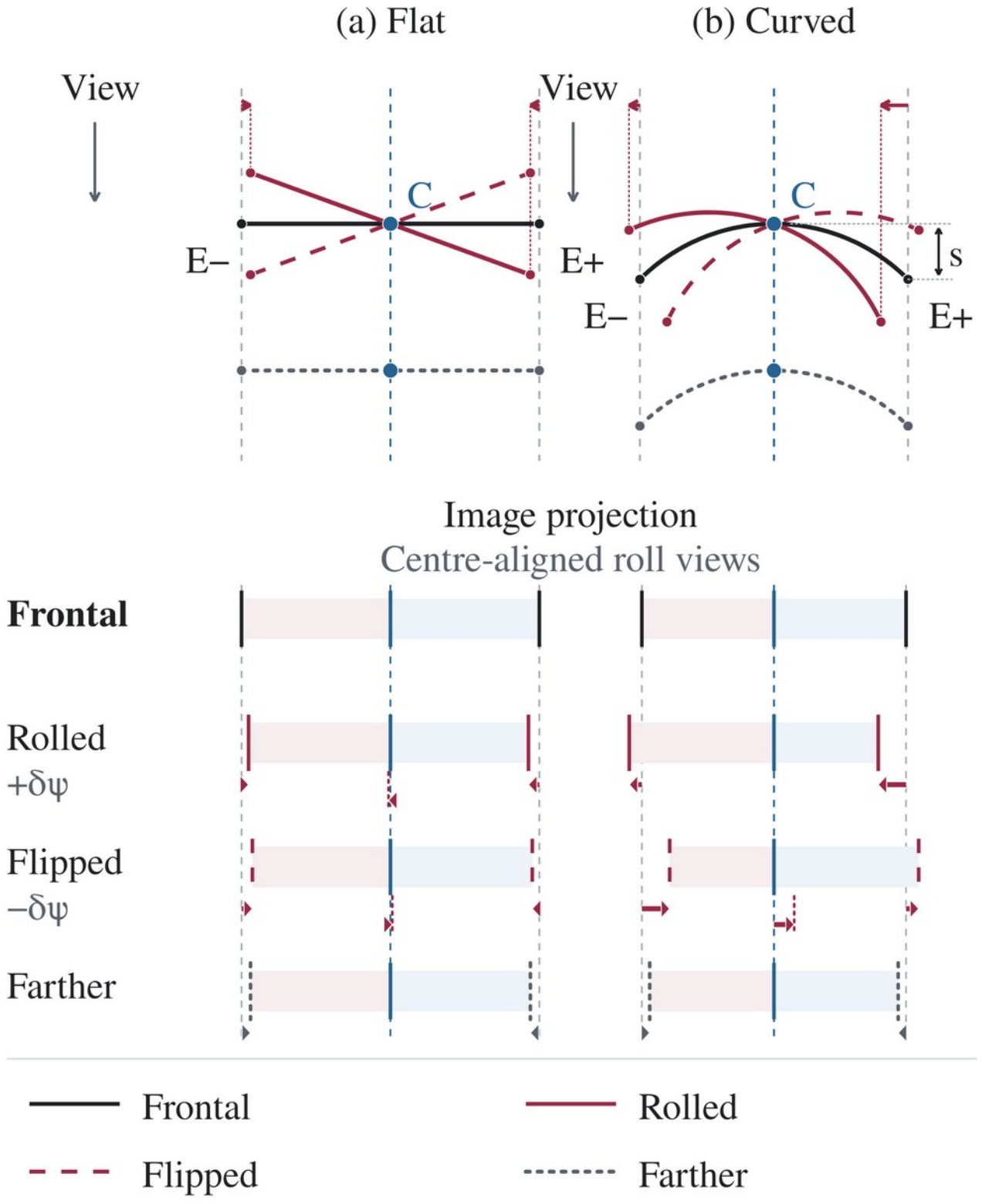

Fig. 3. Projected roll cues for (a) a flat marker and (b) a curved marker. The centre line is aligned in the roll comparison. Grey guides retain the frontal positions; arrows show edge displacement and the edge-midpoint offset from the blue centre line. Line styles distinguish the frontal, rolled, opposite-roll and farther frontal views. Curvature gives a stronger relative roll cue near the frontal pose.

## 4.3. Centre line and projected axis direction

The centre line is the longest continuous feature in the DCS pattern. The colour contrast makes the boundary between the two stripes easy to identify. Its value is not limited to locating the marker. A straight line fitted to many boundary observations estimates the projected shaft direction by combining information across the full marker length. Individual dot centres provide fewer observations, even when each dot is located accurately. The two feature types therefore have complementary roles: the line constrains direction in the image, while identified points supply correspondence and spacing along the shaft for 6-DOF pose estimation.

For Nb independent boundary observations spread uniformly over a projected length L, let σ be the standard deviation of each observation normal to the line. A small-noise approximation gives the following standard deviation of the fitted line direction, in radians:

$$\sigma_{dir} \approx \sigma \sqrt{12/N_b}/L \,(\text{radians}), \quad (4)$$

The relation shows why a long boundary is useful: direction uncertainty falls as projected length increases and as more independent measurements contribute. Figure 4 illustrates this direction uncertainty separately from the roll geometry. The independence assumption matters: neighbouring image pixels can have correlated errors, and sampling a fitted line more densely does not create new independent observations. The relation describes the direction of a projected line, not the uncertainty of the full three-dimensional pose.

A line observation constrains motion across the line but does not locate a point along it. The printed ends or identifiable dots are therefore still needed to establish the axial extent. The centre line and curvature contribute different information: the former supplies a long directional constraint, while the latter creates depth separation. Neither should be treated as a replacement for all other features.

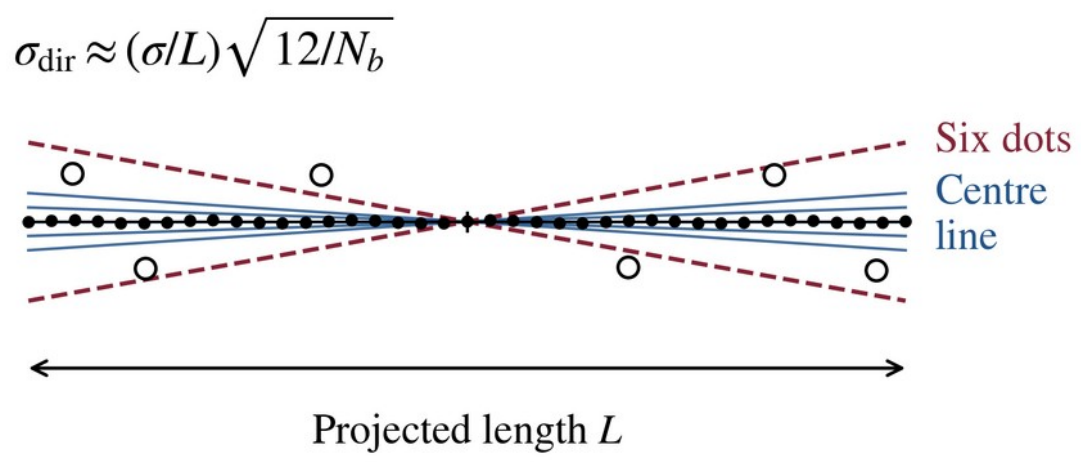

Fig. 4. Direction uncertainty of a fitted centre line. Small black points represent independent boundary observations. The six large open circles represent the inner dot centres, after accounting for their known stripe offsets. Blue lines illustrate centre-line estimates; burgundy dashed lines illustrate estimates based only on the six dot centres. The spreads are schematic and refer to in-image direction, not roll or total pose error.

## 4.4. Comparison with a stereo baseline

Curvature and stereo vision make depth changes observable in different ways. For a flat printed marker, a small roll moves a point at lateral distance h in depth by approximately h δψ. A stereo pair with baseline B converts this into a disparity change of approximately $(fB/\bar{Z}^2)h$ δψ. For a curved printed marker in one view, the relative lateral motion due to the sagitta produces an image change of approximately $(f/\bar{Z})s$ δψ. Equating these two sensitivities gives

$$B_{eq} = s\bar{Z}/h, \quad (5)$$

Here h is the half-width of the corresponding flat print, rather than the projected half-span of the curved band. With h = 6 mm, s ≈ 2.24 mm and $\bar{Z}$ = 150 mm, the equivalent baseline is about 56 mm. Figure 5 compares the two mechanisms. This relation is a local equivalence in roll sensitivity. Since baseline direction also matters, a full pose calculation is needed to determine a baseline that gives equal roll precision.

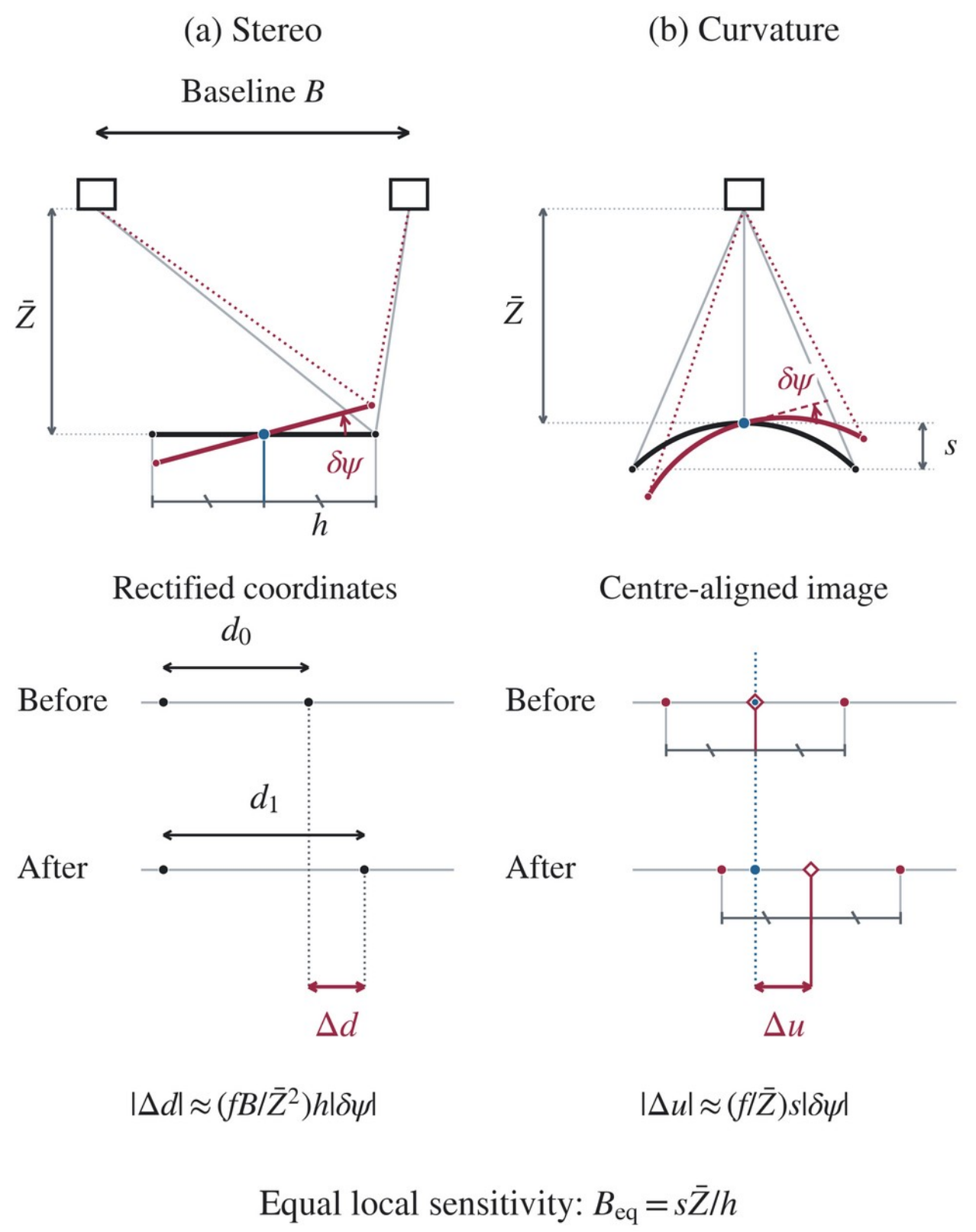

Fig. 5. Local equivalence of roll sensitivity. Black and red marker sections show the initial and rolled poses; dotted red rays connect the cameras to the rolled endpoints. (a) The two points in each row are observations of the same marker feature in the two cameras. Their separation is the stereo disparity, $d_0$ before roll and $d_1$ after roll; $\Delta d = d_1 - d_0$. The rows are aligned at one observation. (b) $\Delta u$ is the change in the edge midpoint relative to the blue centre line. Matching ticks mark equal half-spans; open diamonds identify the midpoint. The coordinate diagrams are schematic. Equal first-order sensitivity does not imply equal pose precision.

## 5. Pose estimation

Figure 6 summarises pose estimation from each input image. The detector extracts the centre line and dots, assigns them to the flat or cylindrical model, and obtains an initial pose. The model is then projected into the image so that the printed outline can provide additional measurements. The same principle applies to flat, curved and wrap-around markers, while the available outline differs with their geometry.

Centre-line and dot detection. After lens distortion is corrected, the two stripe colours are separated in hue–saturation space. Their shared boundary forms a narrow region to which a straight line is fitted. The colour order on its two sides and the asymmetric dot layout establish marker orientation and feature order. Endpoints are located at the physical ends of the printed band. Dots are detected inside each stripe using the contrasting colour. This procedure uses the relationship between the two colour regions, rather than relying only on the external silhouette of the object. The colour model is adjusted before operation to the actual print and camera so that printing and imaging differences do not change the intended identity of the stripes.

Initial pose. Detected features are paired with known model coordinates, as in PnP estimation [19]. For a flat print, planar pose candidates are evaluated against the measured points using established PnP methods [20], [3]. For a curved band, the coordinates are calculated on the cylindrical surface. The apparent displacement of a dot centre caused by curved-surface projection is accounted for when forming the correspondence. For a wrap-around marker, the spacing pattern of each dot row identifies the band and places its points in the common cylinder frame. Dots from neighbouring visible bands are then used together. The initial pose is refined by minimising reprojection error. The advantage of the cylindrical model is that the known depth differences enter the pose estimate instead of being discarded by a planar approximation.

Refinement from the printed outline. An initial pose predicts where the marker boundaries should appear. The input image is searched across these predicted boundaries to locate the actual print transitions. For a flat or single curved band, the useful boundaries are the two long outer edges. On a cylinder they are straight surface lines parallel to the shaft. For a wrap-around strip, adjacent bands continue around the circumference, so the two axial ends provide the external edges. These measurements are combined with the dot and centre-line observations to update the pose. The process links recognition and geometry: colour locates the print, the pose predicts its outline, and the observed edges improve the pose.

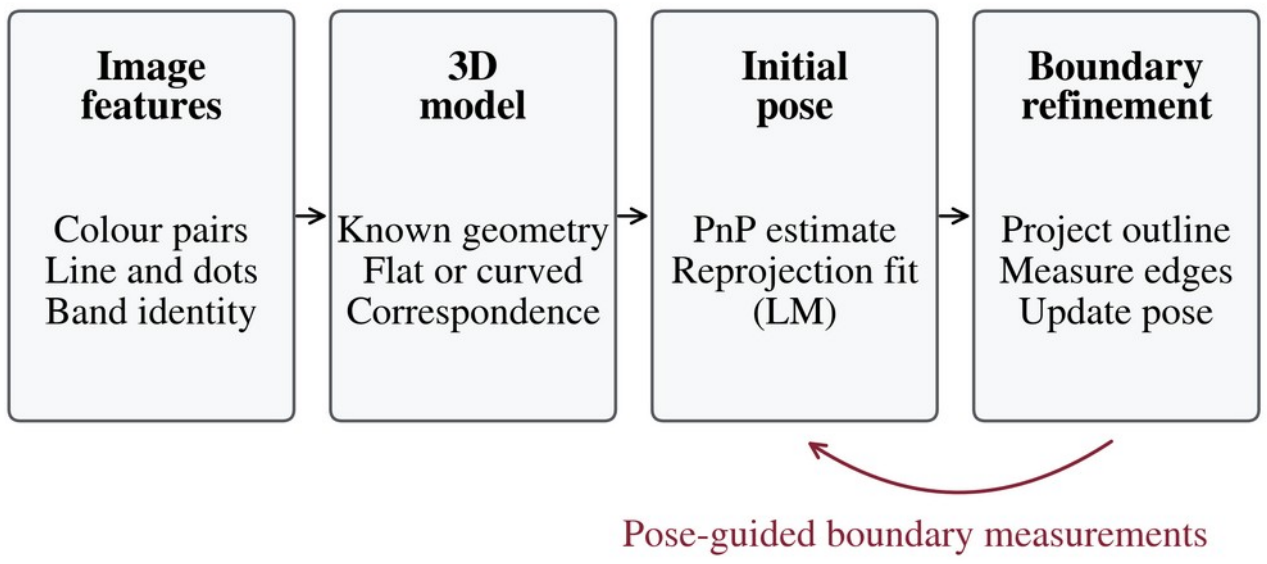

Fig. 6. Pose estimation from an input image. Colour segmentation and band identification establish feature correspondences. PnP gives an initial pose, refined by Levenberg–Marquardt (LM) reprojection fitting. The projected model guides the search for printed boundaries; their measurements update the pose. The cylindrical model retains depth differences between features.

## 6. Simulation study

The simulation study is a virtual experiment that evaluates the proposed geometry under controlled conditions. It separates the information available in the feature arrangement from errors caused by feature detection or printing. We compare roll precision, position and total rotation uncertainty, the contribution of different features, and the effect of a virtual second camera. The results describe the stated observation model and should not be read as measured hardware accuracy.

### 6.1. Simulation method

The marker model was projected with a calibrated pinhole camera using $f = 1458$ px and a $1920 \times 1080$ image. Independent zero-mean Gaussian noise was added to the image observations, and all six pose parameters were re-estimated. Dot centres and centre-line endpoints were treated as two-dimensional observations with 0.3 px standard deviation. Edge observations contributed only their component normal to the projected boundary, with 0.5 px noise for the centre line and 0.3 px for the outer printed edges. Surface orientation determined feature visibility. For the decoded-band comparison, both conditions used the same end boundaries, centre line and identifiable endpoints; only the set of decoded dots differed. This comparison used an effective radius of 7.64 mm. The default condition was a 15 cm working distance and a 5° roll, with 400 trials per condition. Position was evaluated at the cylinder-axis origin for all simulated designs. Monte Carlo errors are reported as root-mean-square (RMS) errors relative to the true pose. Optimisation started at the true pose to measure local noise-limited precision; it did not test recovery from an incorrect initial pose. The corresponding Cramér–Rao covariance bound was calculated from the projection Jacobian and observation noise, with all pose parameters free. Monte Carlo roll RMS errors were within about 10% of the corresponding standard-deviation bounds in the tested cases. This agreement checks the local uncertainty calculation, not detection success or resistance to systematic error. Claude (Anthropic) and ChatGPT/Codex (OpenAI) assisted with implementing the pose-estimation algorithm and developing the simulation and evaluation code. The authors reviewed the code and verified its outputs, including comparisons of Monte Carlo errors with analytical uncertainty bounds.

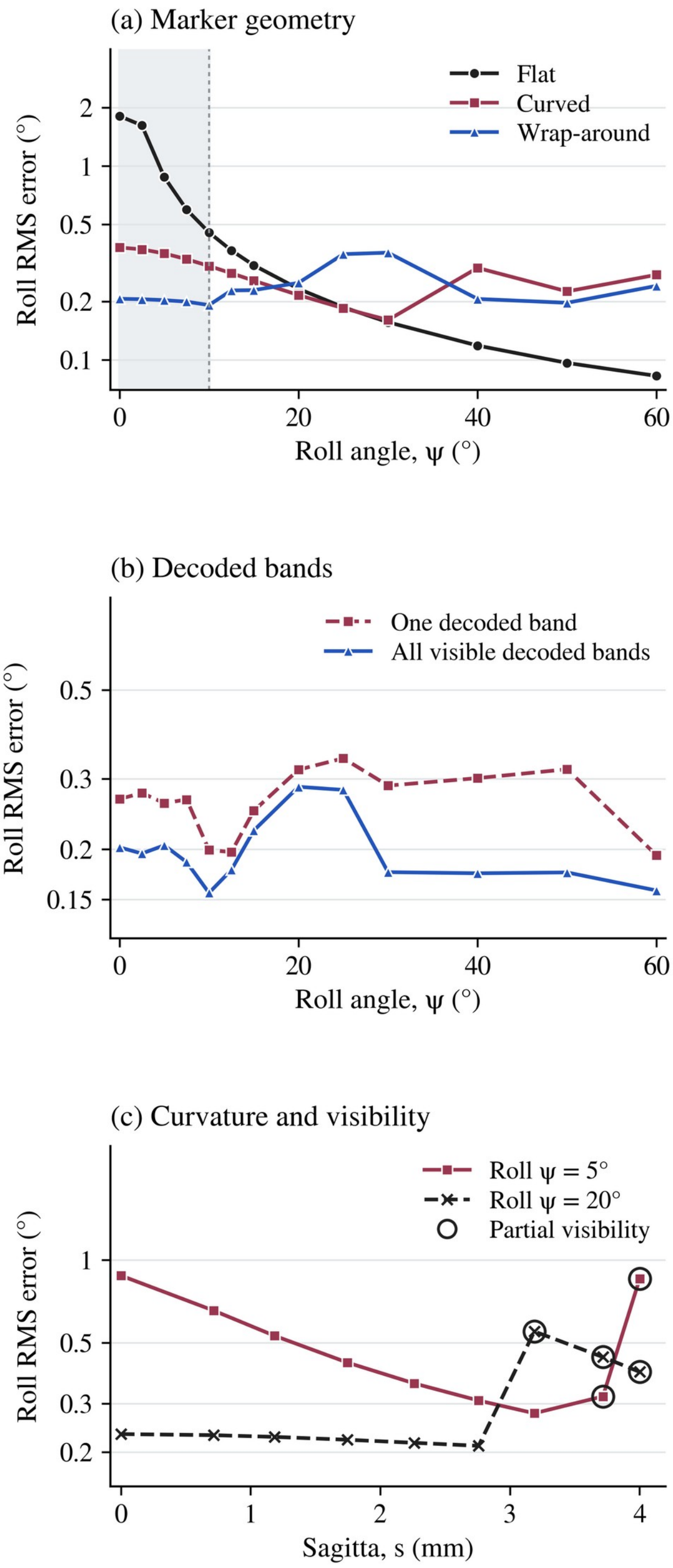

Fig. 7. Simulated roll uncertainty at a 15 cm working distance. (a) Roll RMS error for flat, curved and wrap-around markers; shading marks near-frontal views. (b) A wrap-around marker using one decoded band or all visible decoded bands, with the same end-boundary, centre-line and identifiable endpoint observations in both conditions. (c) Roll RMS error as the sagitta increases; open circles mark partial loss of visibility. All vertical axes use logarithmic scales. Each condition uses 400 Monte Carlo trials.

## 6.2. Roll angle and working distance

The flat marker becomes less precise in roll as it approaches a frontal view. Its simulated roll RMS error increases from 0.45° at 10° roll to 1.81° at zero roll. Over the same interval, the curved marker remains at 0.31–0.38° and the wrap-around marker at 0.19–0.21°. At the frontal pose, the corresponding standard-deviation bounds are 2.00°, 0.38° and 0.21°. The finite flat-marker bound is consistent with the small perspective signal retained by the pinhole model. The advantage is local to the viewing geometry. At larger roll angles, the flat marker provides a stronger width-change signal and can match or exceed the curved designs. In this simulation it matches the wrap-around marker near 19° and the curved marker near 25°. These crossings show why a precision claim must state the viewing range rather than rank marker geometries without conditions.

Table 1. Simulated position RMS and total rotation RMS. Default condition: 15 cm distance, 5° roll and 400 trials. B denotes the stereo baseline; both views contribute to a joint pose estimate. Position is evaluated at the cylinder-axis origin.

| Condition | Flat mm | Flat ° | Curved mm | Curved ° | Wrap mm | Wrap ° |
|---|---|---|---|---|---|---|
| $\bar{Z}$ = 10 cm | 0.052 | 0.56 | 0.056 | 0.34 | 0.031 | 0.17 |
| $\bar{Z}$ = 15 cm | 0.116 | 1.01 | 0.124 | 0.67 | 0.065 | 0.28 |
| $\bar{Z}$ = 20 cm | 0.209 | 1.63 | 0.220 | 1.09 | 0.113 | 0.38 |
| $\bar{Z}$ = 25 cm | 0.335 | 2.44 | 0.353 | 1.59 | 0.174 | 0.49 |
| B = 5.5 mm, along axis | 0.084 | 0.70 | 0.086 | 0.46 | 0.048 | 0.20 |
| B = 22 mm, along axis | 0.072 | 0.60 | 0.075 | 0.35 | 0.042 | 0.18 |
| B = 60 mm, along axis | 0.038 | 0.42 | 0.046 | 0.26 | 0.025 | 0.14 |
| B = 120 mm, along axis | 0.023 | 0.27 | 0.030 | 0.20 | 0.016 | 0.10 |
| B = 5.5 mm, across strip | 0.079 | 0.59 | 0.083 | 0.45 | 0.048 | 0.20 |
| B = 22 mm, across strip | 0.050 | 0.34 | 0.057 | 0.30 | 0.044 | 0.20 |
| B = 60 mm, across strip | 0.023 | 0.16 | 0.029 | 0.16 | 0.036 | 0.21 |
| B = 120 mm, across strip | 0.015 | 0.09 | 0.025 | 0.19 | 0.023 | 0.16 |

## 6.3. Curvature and feature selection

Increasing distance worsens precision for all three geometries. At the default roll angle, the wrap-around design has the smallest total rotation RMS at every simulated distance, followed by the curved and flat designs. At 15 cm, their roll RMS values are 0.20°, 0.35° and 0.88°, respectively. The corresponding axis-direction errors are 0.19°, 0.57° and 0.50°. Thus a single curved band improves roll without improving every rotation component. In particular, it does not improve axis-direction precision over the flat print in this comparison. The wrap-around design improves both components because its decoded features are distributed over more of the circumference.

The benefit of increasing curvature is limited by visibility. For the same 40 × 12 mm print, reducing cylinder radius increases the sagitta and generally strengthens the roll signal while the required features remain visible. At a 5° roll, the roll RMS decreases from 0.88° for the flat print to 0.28° at a 5 mm radius, with less than 0.01 mm change in position error. A smaller radius, however, places more of the band around the hidden side of the object. The design must therefore balance depth separation with the required angular field of view. The present 12 mm band covers about one quarter of the circumference at R = 7.64 mm. A wrap-around design addresses this visibility limit by supplying identifiable features from neighbouring bands. Combining all visible decoded bands, with the same boundary observations, reduced roll RMS error from 0.27° to 0.20° at zero roll and from 0.32° to 0.18° at 50° roll compared with using one band.

### 6.4. Stereo baseline for equal roll precision

We also calculated the stereo baseline at which a flat marker reaches the roll bound of a curved marker observed by one camera. At 15 cm and 5° roll, the required baseline is 14–73 mm for the single curved band and 33–152 mm for the wrap-around design. The lower values correspond to a baseline across the strip and the higher values to one along the shaft. Direction matters because a line observation supplies information across the line, not along it. The simple sensitivity estimate of 56 mm lies within these ranges but does not replace the full calculation. The required baseline increases with working distance and decreases as the flat marker turns away from the frontal pose. A monocular curved marker can therefore provide roll sensitivity comparable to that of a stereo baseline under specified conditions. Curvature is particularly useful where a single camera is preferred or where a large baseline is impractical.

## 7. Discussion

The main finding is a selective benefit of curvature. For a thin marker viewed near the front, depth-separated observations improve roll estimation because they create relative image motion that is difficult to explain by translation alone. The single curved band does not improve all pose components: its simulated position and axis-direction results are similar to those of the flat print under the simulated conditions. The wrap-around design provides a wider feature distribution and performs better across several measures. These distinctions keep the interpretation of the simulation results aligned with the geometric analysis.

Marker choice should follow the expected motion and visible surface area. A single curved band uses a small circumferential area and is suitable when large axial rotation is not required. A wrap-around strip preserves identifiable features over a wider roll range and defines absolute material orientation, at the cost of a larger print and more feature processing. Colour selection must also suit the camera and illumination.

The stereo comparison describes a design trade-off, not a replacement claim. A second camera can provide stronger geometric constraints, especially with a sufficiently large baseline across the strip. It also requires another image channel, synchronisation and an extrinsic calibration. Curvature provides useful depth separation within a single view by using the object's existing shape. The equivalent baseline relation helps interpret that information, while the full simulation shows how strongly the answer depends on feature type, viewing distance and baseline direction. Small integrated stereo systems should therefore be assessed at their actual baseline rather than assumed to be equivalent to an external stereo rig.

The contribution also differs from a general claim of novelty for cylindrical markers or model-based refinement. Both have established precedents [7], [14], [15], [16], [8]. However, the present work connects a specific thin colour-stripe design to a quantitative analysis of roll information. It explains why the outer edges matter for one curved band, why decoded neighbouring bands provide stronger geometry, and how the gain compares with an added camera. The relevant relations are derived here and tested in simulation.

Several limits remain. The uncertainty analysis assumes independent Gaussian image noise, correct correspondences and a correct surface model. Real boundaries can have correlated errors and systematic shifts. The simulations evaluate local noise-limited precision and do not establish detection reliability, real-time tracking or hardware accuracy. Finally, independent calibration, controlled motion and application-specific validation are required before drawing stronger claims about accuracy or application performance.

## 8. Conclusion

We proposed a dual-colour-stripe marker that uses curvature of known cylindrical geometry for monocular pose measurement. The long centre line supports axis-direction estimation, while dots and printed boundaries at different depths improve the constraints on pose. The analysis identifies the sagitta as a source of first-order roll information near a frontal view and relates its image sensitivity to that of a stereo baseline. A separate simulation study quantifies the effects of radius, working distance, roll and feature selection. It also shows that the benefit is component-specific and depends on visibility.

These analytical and simulation results support curvature as a useful design resource for thin markers observed by one camera.

## Data availability statement

The simulation data supporting the findings of this study are available from the corresponding author upon reasonable request. The data are not publicly available to protect intellectual property associated with potential future patent applications.

## Acknowledgements

This work was supported by the National Research Foundation of Korea (NRF) grant funded by the Korea government (MSIT) (No. RS2024-00355535).

Generative AI disclosure. The authors determined the scientific content, manuscript organisation, figure content and layouts. Claude (Anthropic) and ChatGPT/Codex (OpenAI) assisted with text drafting and revision, algorithm implementation, simulation and evaluation code, and figure rendering. The authors reviewed all AI-assisted code, verified the computational outputs and revised the manuscript and figures. Plotted values were obtained from the numerical simulations and verified by the authors. The authors take full responsibility for the final manuscript.

Conflict of interest: The authors declare no conflicts of interest.